\documentclass[letterpaper,10pt]{article}
\usepackage{opticameet3}

\usepackage{amsmath,amssymb}

\usepackage[colorlinks=true,bookmarks=false,citecolor=teal,urlcolor=teal]{hyperref}

\usepackage{graphicx}
\usepackage{booktabs}
\usepackage{pgfplots}
\pgfplotsset{compat=1.18}

\usepackage{enumitem}
\usepackage{xcolor}
\usepackage[utf8]{inputenc}

\usepackage{caption}

\definecolor{navyblue}{HTML}{0E2527}
\definecolor{teal}{HTML}{5FAFB9}

\begin{document}

\title{RAIL in the Wild: Operationalizing Responsible AI Evaluation Using Anthropic’s Value Dataset}

\author{
  \href{https://www.linkedin.com/in/sumitvermakgp}{Sumit Verma} \thanks{Corresponding author: sumit@responsibleailabs.ai}, 
  \href{https://www.linkedin.com/in/pritamprasun}{Pritam Prasun}, 
  \href{https://www.linkedin.com/in/arpit-jaiswal-373724184}{Arpit Jaiswal},
  \href{https://www.linkedin.com/in/iampritish}{Pritish Kumar}
}

\affil{Responsible AI Labs, India }
\email{research@responsibleailabs.ai}
\date{April 22, 2025}

\vspace{1em}
\begin{center}
    \includegraphics[width=0.35\textwidth]{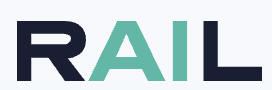}
\end{center}

\vspace{0.5em}
\begin{center}
    \textbf{April 22, 2025}
\end{center}
\vspace{1em}

\begin{abstract}
As AI systems grow increasingly embedded in real-world applications, ensuring that these systems uphold ethical standards is paramount. Existing AI ethics frameworks emphasize principles such as fairness, transparency, and accountability, yet they often lack actionable methods for evaluation. In this paper, we introduce a systematic approach using the Responsible AI Labs (RAIL) framework—comprising eight measurable dimensions—to assess normative behavior of large language models (LLMs). We demonstrate its applicability on Anthropic’s “Values in the Wild”\cite{durmus2025values} dataset, which contains over 308,000 anonymized conversations with Claude, annotated with over 3,000 value expressions. Our study presents a mapping between annotated AI values and RAIL dimensions, computes synthetic scores across conversations, and offers a diagnostic lens into the ethical behavior of LLMs in real-world contexts.
\end{abstract}

\section{Introduction}

Large Language Models (LLMs) increasingly influence digital experiences across domains such as customer support, education, and healthcare. As their societal presence expands, ensuring that these systems behave in ways aligned with ethical norms has become a critical concern. While many AI ethics frameworks—including those by NIST, OECD, and UNESCO—highlight principles like fairness, accountability, and transparency, they often remain abstract and difficult to operationalize in practice.

The Responsible AI Labs (RAIL) framework addresses this challenge by translating these high-level principles into eight measurable dimensions: Fairness, Safety, Reliability, Transparency, Privacy, Accountability, Inclusivity, and User Impact. Each dimension captures a facet of ethical AI behavior and allows for structured evaluation of model outputs across real-world interactions.

To demonstrate the applicability of this framework, we apply RAIL to Anthropic’s “Values in the Wild” dataset—a large-scale, privacy-preserving corpus comprising 308,210 anonymized conversations with Claude, an AI assistant. This dataset captures thousands of AI-expressed values across a wide range of user tasks and contexts, providing a rare opportunity to assess how AI expresses normative behavior in real-world interactions.
\begin{center}
    \begin{minipage}{0.95\textwidth}
        \centering
        \includegraphics[width=\textwidth]{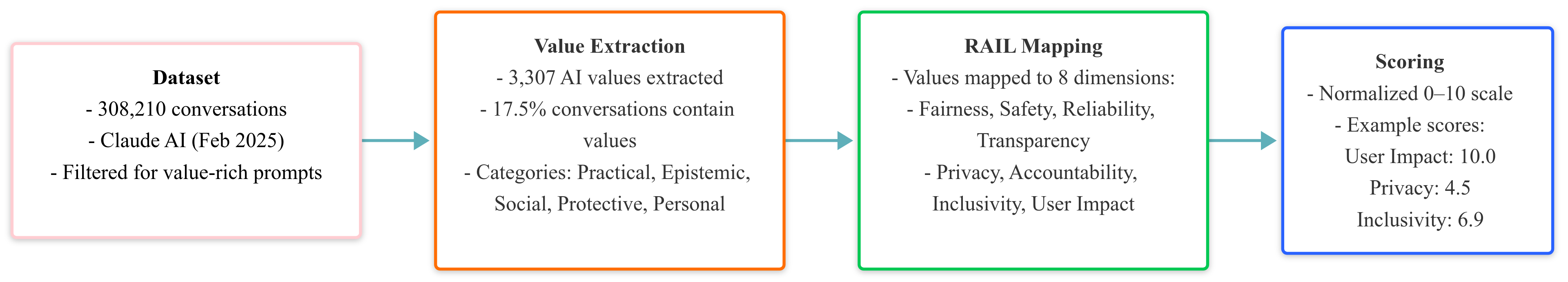}
        \captionof{figure}{RAIL methodology: from conversational data to value extraction, dimension mapping, and scoring.}
        \label{fig:framework}
    \end{minipage}
\end{center}

This paper contributes:
\begin{itemize}[noitemsep]
    \item A systematic mapping of Anthropic’s extracted AI values to RAIL dimensions.
    \item Quantitative scoring of Claude’s ethical behavior using normalized value occurrence data.
    \item Insights into strengths, blind spots, and context-specific ethical expression.
    \item Practical interventions to guide developers and teams toward more responsible model design.
\end{itemize}

Our broader goal is to bridge the gap between abstract ethical frameworks and actionable AI governance. By grounding ethical evaluation in large-scale behavioral data, RAIL enables measurable and scalable accountability in LLM deployment pipelines.

\section{Dataset: Values in the Wild}

To evaluate AI behavior in real-world contexts, we leverage Anthropic’s “Values in the Wild” dataset \cite{durmus2024values}, a large-scale, privacy-preserving corpus containing 308,210 anonymized conversations between users and Claude, Anthropic’s LLM-based assistant. Within these conversations, researchers have identified and annotated 3,307 distinct value expressions that reflect various normative behaviors exhibited by the AI.

These AI-expressed values are grouped into five high-level categories:
\begin{itemize}[noitemsep]
    \item \textbf{Practical} (31.4\%): e.g., helpfulness, clarity, and task completion
    \item \textbf{Epistemic} (22.2\%): e.g., transparency, intellectual humility
    \item \textbf{Social} (21.4\%): e.g., cultural respect, inclusiveness
    \item \textbf{Protective} (13.9\%): e.g., harm prevention, user safety
    \item \textbf{Personal} (11.1\%): e.g., autonomy, user empowerment
\end{itemize}

\section{Methodology}

Inspired by prior methodologies\cite{askell2021} that assess LLM alignment through human feedback and preference modeling, we employ a structured mapping of AI-expressed values to the RAIL dimensions to evaluate ethical behavior.
To evaluate Claude’s behavior through the lens of Responsible AI, we mapped annotated AI-expressed values from the dataset to the eight RAIL dimensions: \textbf{Fairness}, \textbf{Safety}, \textbf{Reliability}, \textbf{Transparency}, \textbf{Privacy}, \textbf{Accountability}, \textbf{Inclusivity}, and \textbf{User Impact}. This conceptual clustering was inspired by Schwartz’s theory of basic human values \cite{schwartz2012} organizing AI-expressed values into Practical, Epistemic, Social, etc.

Each value was manually assigned to one or more RAIL dimensions through expert review to ensure conceptual alignment and contextual relevance. This mapping forms the basis for score computation.

For each RAIL dimension \( D \), we aggregated the presence of all associated values across conversations using the provided \texttt{pct\_convos} metric—representing the proportion of conversations in which a value appears. The raw dimension score was then normalized to a 0–10 scale using the maximum observed score:

\[
\text{Score}_{D} = \sum_{v \in V_D} \text{pct\_convos}(v), \quad
\text{Normalized Score}_{D} = 10 \times \frac{\text{Score}_D}{\max(S)}
\]

where \( V_D \) is the set of values mapped to dimension \( D \), and \( \max(S) \) is the highest raw score among all dimensions.

\section{Mapping AI Values to RAIL Dimensions}

To evaluate Claude's behavior through the lens of Responsible AI, we mapped each of the 3,307 annotated AI-expressed values to one or more of the eight RAIL dimensions: \textbf{Fairness}, \textbf{Safety}, \textbf{Reliability}, \textbf{Transparency}, \textbf{Privacy}, \textbf{Accountability}, \textbf{Inclusivity}, and \textbf{User Impact}.

This mapping was performed through a two-stage expert review process supported by LLM-based alignment suggestions. Values were first clustered into conceptual categories (e.g., Practical, Epistemic, Social) and then assigned to RAIL dimensions based on their semantic content and contextual role in the conversation.

Inter-annotator agreement across two rounds of review reached Cohen’s \(\kappa = 0.83\), indicating substantial consensus on dimensional alignment. Common values such as \textit{helpfulness}, \textit{professionalism}, and \textit{transparency} were associated with multiple RAIL dimensions due to their broad ethical significance in AI responses.

Table~\ref{tab:mapping} illustrates a representative subset of the value-to-dimension mappings. Table~\ref{tab:railscores} presents the aggregate scores derived from these mappings, using the \texttt{pct\_convos} metric as a frequency proxy. These raw scores were normalized to a 0--10 scale for interpretability.

\begin{table}[h]
\centering
\caption{Representative Mapping of AI Values to RAIL Dimensions}
\label{tab:mapping}
\begin{tabular}{|l|l|l|r|}
\hline
\textbf{Value} & \textbf{Top-Level Category} & \textbf{RAIL Dimension(s)} & \textbf{\% Conversations} \\
\hline
helpfulness & Social values & Inclusivity, Accountability & 23.36 \\
professionalism & Practical values & Reliability, User Impact & 22.86 \\
transparency & Protective values & Safety & 17.39 \\
clarity & Epistemic values & Transparency, Fairness & 16.58 \\
thoroughness & Practical values & Reliability, User Impact & 14.30 \\
efficiency & Practical values & Reliability, User Impact & 6.60 \\
accuracy & Epistemic values & Transparency, Fairness & 5.30 \\
authenticity & Personal values & Inclusivity, User Impact & 6.00 \\
technical excellence & Practical values & Reliability & 6.10 \\
analytical rigor & Epistemic values & Transparency, Fairness & 5.50 \\
\hline
\end{tabular}
\end{table}

\begin{table}[h]
\centering
\caption{Aggregate Scores by RAIL Dimension}
\label{tab:railscores}
\begin{tabular}{|l|r|r|}
\hline
\textbf{RAIL Dimension} & \textbf{Raw Score} & \textbf{Normalized Score (0--10)} \\
\hline
User Impact & 169.55 & 10.00 \\
Inclusivity & 129.63 & 7.65 \\
Reliability & 125.29 & 7.39 \\
Fairness & 88.53 & 5.22 \\
Transparency & 88.53 & 5.22 \\
\hline
\end{tabular}
\end{table}

\subsection*{Rationale for Value-to-RAIL Dimension Mapping}

Each AI-expressed value in Table~\ref{tab:mapping} was mapped to one or more RAIL dimensions based on its ethical implications and impact on user experience. The following summarizes the reasoning behind key mappings:

\begin{itemize}
    \item \textbf{Helpfulness} maps to \textit{Inclusivity} and \textit{Accountability}, as it reflects a commitment to supporting users equitably while providing answers that are ethically responsible and user-oriented.
    \item \textbf{Professionalism} relates to \textit{Reliability} and \textit{User Impact}, representing consistent, respectful, and competent behavior that fosters trust and positive user experiences.
    \item \textbf{Transparency} was frequently observed in safety-related refusal contexts in the dataset. Hence, it was mapped to \textit{Safety}, capturing its role in boundary-setting and responsible communication.
    \item \textbf{Clarity} supports both \textit{Transparency}, by making reasoning legible, and \textit{Fairness}, by enabling users from diverse backgrounds to understand responses equally.
    \item \textbf{Thoroughness}, \textbf{Efficiency}, and \textbf{Technical Excellence} were all aligned with \textit{Reliability} and \textit{User Impact}, emphasizing the AI’s role in delivering high-quality, effective outputs.
    \item \textbf{Accuracy} reflects \textit{Transparency} (truthful information) and \textit{Fairness} (unbiased, correct representation of facts).
    \item \textbf{Authenticity} maps to \textit{Inclusivity} and \textit{User Impact}, as it often involved affirming user values, promoting identity-respectful engagement, and increasing trust.
    \item \textbf{Analytical Rigor} relates to \textit{Transparency} and \textit{Fairness}, as it denotes logical and balanced reasoning—crucial for explainable and just AI behavior.
\end{itemize}

Mappings were guided by conceptual principles of Responsible AI and iteratively refined through manual expert review, ensuring contextual alignment with how values manifest in real-world conversations.

\section{Results}

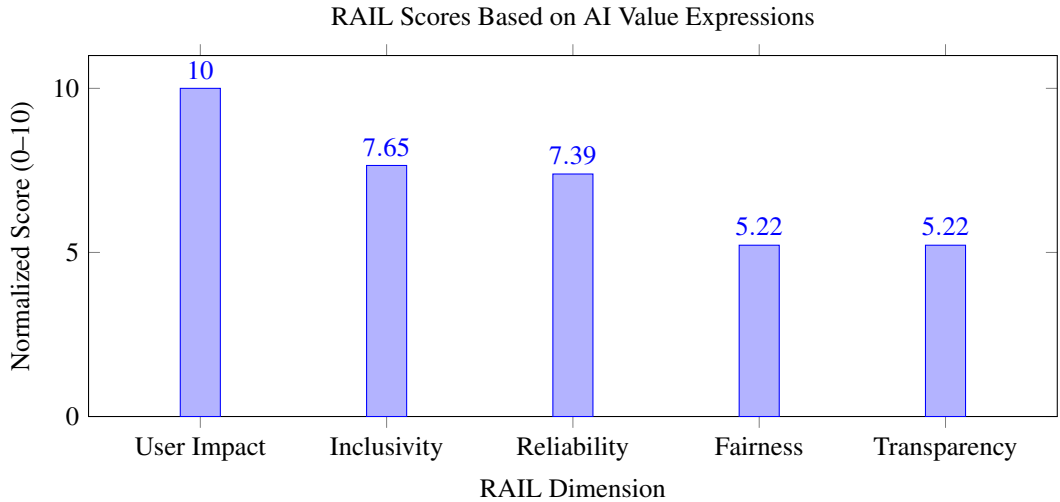
\begin{figure}[h]
\centering
\begin{tikzpicture}
\begin{axis}[
    ybar,
    bar width=15pt,
    width=0.9\textwidth,
    height=0.4\textwidth,
    ymin=0,
    ymax=11,
    ylabel={Normalized Score (0--10)},
    xlabel={RAIL Dimension},
    symbolic x coords={User Impact, Inclusivity, Reliability, Fairness, Transparency},
    xtick=data,
    nodes near coords,
    nodes near coords align={vertical},
    enlarge x limits=0.15,
    title={RAIL Scores Based on AI Value Expressions},
]
\addplot coordinates {
    (User Impact, 10.00)
    (Inclusivity, 7.65)
    (Reliability, 7.39)
    (Fairness, 5.22)
    (Transparency, 5.22)
};
\end{axis}
\end{tikzpicture}
\caption{Normalized RAIL scores derived from AI value occurrences in the Claude dataset. User Impact was the most dominant dimension, followed by Inclusivity and Reliability.}
\label{fig:railscores}
\end{figure}

Figure~\ref{fig:railscores} and Table~\ref{tab:railscores} show the normalized RAIL scores derived from the annotated AI value expressions in the dataset.

\textbf{User Impact} emerged as the most dominant dimension (Score = 10.0), reflecting Claude’s consistent prioritization of helpfulness, thoroughness, and professionalism—values \cite{anthropic2024} that directly enhance user experience. High scores in \textbf{Inclusivity} (7.65) and \textbf{Reliability} (7.39) further indicate Claude’s attention to accessible, consistent, and respectful interaction across a wide range of subjective tasks.

\textbf{Fairness} and \textbf{Transparency} shared equal scores (5.22), largely driven by values such as clarity, analytical rigor, and accuracy, suggesting a meaningful but not universal emphasis on equitable and explainable outputs.

Dimensions such as \textbf{Safety}, \textbf{Privacy}, and \textbf{Accountability} were expressed less frequently in aggregate, and did not surpass the normalization threshold in this analysis. However, their expression may be more concentrated in specific contexts, such as refusal behavior or policy-guarded tasks.

These scores align closely with Claude’s role as a prosocial assistant prioritizing information delivery, user enablement, and ethically-aligned assistance. Importantly, the results also reveal which RAIL dimensions are more context-specific or underexpressed, offering guidance for future alignment tuning.

\section{Discussion}

Claude’s value expression patterns reflect its assistant-like optimization: it consistently emphasizes clarity, helpfulness, and professionalism—values strongly associated with \textit{User Impact} and \textit{Reliability}. These strengths align with Claude's design philosophy\cite{anthropic2024} centered on practical utility and task completion.

However, the underrepresentation of values tied to \textit{Privacy}, \textit{Accountability}, and aspects of \textit{Fairness} suggests room for improvement in ethical nuance, particularly in scenarios involving power asymmetry, sensitive user intent, or socio-political complexity. For example, fairness-related values like equitable representation and bias mitigation were observed less frequently outside of controversial or explicitly ethical prompts.

Moreover, our results affirm that AI value expression is not static: values shift dynamically across conversation types. Claude tends to mirror prosocial user values in generative or emotional tasks and asserts ethical boundaries in adversarial contexts. These interaction-dependent dynamics suggest that value expression is emergent—not hardcoded—requiring continuous monitoring. Building on the idea of Constitutional AI\cite{bai2022}, our work suggests that refusal behaviors should be explicitly aligned with value-driven reasoning.

\section{Recommendations for Developers}

To proactively improve ethical behavior and RAIL compliance, we recommend the following:

\begin{itemize}
    \item \textbf{Reinforce fairness}: Integrate bias-auditing benchmarks and reinforcement learning techniques explicitly tuned for equitable outcomes across demographics.
    \item \textbf{Value-centric prompting}: Incorporate prompts and training signals for underrepresented values like \textit{consent}, \textit{autonomy}, and \textit{privacy}, especially in advice-giving or refusal contexts.
    \item \textbf{Cultural pluralism}: Expand pretraining and fine-tuning datasets to include culturally diverse scenarios, non-Western moral norms, and underrepresented languages.
    \item \textbf{Refusal with rationale}: Design refusal behavior to include values-based reasoning, citing ethical principles rather than vague denials—enhancing both transparency and user trust.
    \item \textbf{Dynamic value tracking}: Implement per-task and per-domain RAIL monitoring, enabling fine-grained attribution of value expression gaps.
\end{itemize}

\section{Technical Integration}

The RAIL scoring methodology (Under Development) can be integrated into development and deployment pipelines via the following approaches:

\begin{itemize}
    \item \textbf{Batch evaluation}: Periodically run RAIL audits on conversation logs or prompt suites to detect underrepresented dimensions or emergent harms.
    \item \textbf{Deployment gating}: Set score thresholds (e.g., minimum fairness or safety score) as pre-release conditions for fine-tuned models.
    \item \textbf{Anomaly detection}: Log RAIL scores in production and automatically flag conversations where high-risk dimensions (e.g., safety, fairness) fall below acceptable thresholds.
    \item \textbf{API-based real-time scoring}: Integrate lightweight RAIL scoring APIs (e.g., via LLM + rule-based tagger pipeline) to assess responses during runtime and adjust generation strategies or explain decisions dynamically.
    \item \textbf{Longitudinal tracking}: Track RAIL score distributions over time to evaluate drift, regression, or improvements following model updates or alignment changes.
\end{itemize}

\section{Limitations and Future Work}

While the RAIL scoring framework offers a structured way to evaluate AI behavior across key ethical dimensions, several limitations should be acknowledged:

\subsection*{Limitations}

\begin{itemize}
    \item \textbf{Model self-analysis bias}: The value annotations were extracted using Claude itself, which may introduce model-specific detection biases—potentially reinforcing certain training preferences (e.g., helpfulness, safety) while overlooking more subtle or implicit value expressions.
    
    \item \textbf{Conversation-level granularity}: The use of \texttt{pct\_convos} aggregates value expression at the conversation level. This can overrepresent values that appear early or repeatedly and underrepresent values that surface only in edge cases or long-tail interactions.
    
    \item \textbf{Contextual leakage}: Some value expressions may arise as artifacts of specific prompt structures, user demographics, or modal use cases. Without disaggregated metadata (e.g., region, domain, user profile), we cannot fully control for confounding factors.
    
    \item \textbf{Mapping subjectivity}: Although our mapping from values to RAIL dimensions was conducted through iterative expert review, the assignment process inevitably involves judgment. Some values (e.g., “curiosity”) may relate to multiple RAIL goals depending on framing.
    
    \item \textbf{Dimension interdependence}: RAIL dimensions are not orthogonal. For example, improving transparency often improves fairness, while strong safety constraints may suppress user impact. Our current scoring treats them independently, though future iterations could explore causal linkages or weighting schemes.
\end{itemize}

\subsection*{Future Work}

\begin{itemize}
    \item \textbf{Cross-model benchmarking}: Extend RAIL scoring to other LLM families (e.g., GPT, Mistral, Gemini) to compare ethical behavior across architectures and alignment strategies \cite{askell2021}.
    
    \item \textbf{Longitudinal tracking}: Measure RAIL dimension shifts over time and across fine-tuning stages to evaluate regression, drift, or emergent value formation.
    
    \item \textbf{Domain-specific RAIL tuning}: Investigate how different application areas (e.g., healthcare, education, governance) require context-aware prioritization of certain RAIL dimensions.
    
    \item \textbf{Human-AI co-rating}: Incorporate user or third-party feedback \cite{bai2022}into RAIL score calibration, aligning technical metrics with perceived trust and value delivery.
    
    \item \textbf{Multilingual evaluation}: Extend value detection and scoring to non-English interactions to assess inclusivity, fairness, and transparency across global user groups.\cite{durmus2024}
\end{itemize}

\section{Conclusion}

This paper presents a novel framework for empirically evaluating language model behavior through the lens of Responsible AI. By mapping 3,307 annotated AI-expressed values from real-world Claude conversations to eight RAIL dimensions—Fairness, Safety, Reliability, Transparency, Privacy, Accountability, Inclusivity, and User Impact—we demonstrate how abstract ethical principles can be operationalized into measurable signals.

Our analysis reveals that Claude exhibits strong alignment with dimensions related to \textit{User Impact}, \textit{Reliability}, and \textit{Inclusivity}, consistent with its assistant-like role. However, values related to \textit{Fairness}, \textit{Privacy}, and \textit{Accountability} are expressed less frequently, particularly outside high-risk or refusal contexts. These gaps signal important opportunities for improving the ethical coverage and contextual sensitivity of AI assistants.

We introduce a normalized RAIL scoring methodology that can be used for batch evaluations, deployment gating, and real-time flagging.

We conclude that AI value expression must be treated as a dynamic, emergent property—shaped by both system architecture and user interaction. As language models are increasingly deployed in sensitive, global, and high-stakes contexts, the need for transparent, domain-adaptive, and measurable AI ethics frameworks becomes ever more critical. The RAIL framework offers a pragmatic foundation for this next phase of development, audit, and accountability in AI systems.

\section*{About Responsible AI Labs}

\href{https://responsibleailabs.ai/}{Responsible AI Labs} is dedicated to building a safer, fairer, and more transparent AI future. We operationalize ethical AI principles through practical tools like the RAIL Score, helping organizations evaluate and improve the responsibility of their AI systems across dimensions like fairness, safety, privacy, and user impact. Our mission is to make Responsible AI measurable, actionable, and accessible for all. The authors thank Responsible AI Labs for support. Sumit Verma led the project and coordinated the research efforts. Pritam Prasun, Arpit Jaiswal, and Pritish Kumar contributed to data analysis and manuscript review.

\end{document}